\documentclass{article}

\PassOptionsToPackage{numbers, compress}{natbib}

\usepackage[preprint]{neurips_2025}

\usepackage[utf8]{inputenc}
\usepackage[T1]{fontenc}
\usepackage{hyperref}
\usepackage{url}
\usepackage{booktabs}
\usepackage{amsfonts}
\usepackage{nicefrac}
\usepackage{microtype}
\usepackage{xcolor}
\usepackage{multirow}
\usepackage{amsmath,amssymb}
\usepackage{enumitem}
\usepackage{xspace}
\usepackage{graphicx}
\usepackage{array}
\usepackage{float}
\usepackage{booktabs}
\usepackage{multirow}
\usepackage{longtable}
\usepackage[table]{xcolor}
\usepackage{array}
\usepackage{colortbl}
\usepackage{booktabs}
\usepackage{multirow}
\usepackage{longtable}
\usepackage[table]{xcolor}
\usepackage{array}
\usepackage{colortbl}
\usepackage[normalem]{ulem}
\usepackage{caption}
\usepackage{makecell}
\definecolor{bestlavender}{RGB}{223,221,247}   
\definecolor{gaingreen}{RGB}{0,170,0}          

\newcommand{\bestcell}[1]{\cellcolor{bestlavender}\textbf{#1}}

\newcolumntype{G}{>{\columncolor{stagegreen}}c}

\newcommand{\method}{AIM\xspace}

\newcommand{\mot}{mixture-of-transformers\xspace}

\title{AIM: Intent-Aware Unified world action Modeling with Spatial Value Maps}

\author{%
  \textrm{Liaoyuan Fan}$^{1,2}$, \textrm{Zetian Xu}$^{1,2}$, \textrm{Chen Cao}$^{1,2}$, \textrm{Wenyao Zhang}$^{3}$, \textrm{Mingqi Yuan}$^{1,2}$, \textrm{Jiayu Chen}$^{1,2}$ \\[0.8em]
  $^{1}$INFIFORCE Intelligent Technology Co., Ltd., Hangzhou, China \\
  $^{2}$The University of Hong Kong, Hong Kong SAR, China \\
  $^{3}$Shanghai Jiao Tong University, Shanghai, China  \\[0.8em]
}

\begin{document}
\maketitle

\begin{abstract}
Pretrained video generation models provide strong priors for robot control, but existing unified world action models still struggle to decode reliable actions without substantial robot-specific training. We attribute this limitation to a structural mismatch: while video models capture how scenes evolve, action generation requires explicit reasoning about where to interact and the underlying manipulation intent.
We introduce \textbf{\method}, an intent-aware unified world action model that bridges this gap via an explicit spatial interface. Instead of decoding actions directly from future visual representations, \method predicts an aligned spatial value map that encodes task-relevant interaction structure, enabling a control-oriented abstraction of future dynamics. Built on a pretrained video generation model, \method jointly models future observations and value maps within a shared mixture-of-transformers architecture. It employs intent-causal attention to route future information to the action branch exclusively through the value representation. We further propose a self-distillation reinforcement learning stage that freezes the video and value branches and optimizes only the action head using dense rewards derived from projected value-map responses together with sparse task-level signals. To support training and evaluation, we construct a simulation dataset of 30K manipulation trajectories with synchronized multi-view observations, actions, and value-map annotations. Experiments on RoboTwin 2.0 benchmark show that \method achieves a 94.0\% average success rate, significantly outperforming prior unified world action baselines. Notably, the improvement is more pronounced in long-horizon and contact-sensitive manipulation tasks, demonstrating the effectiveness of explicit spatial-intent modeling as a bridge between visual world modeling and robot control.
\end{abstract}

\section{Introduction}
Pretrained video generation models~\cite{dreamzero,wan2025wan,hu2024vpp,liang2025videogenerators} have emerged as a promising foundation for robot learning because they capture rich visual dynamics, object interactions, and multi-step temporal structure from large-scale video data. In parallel, Vision-Language-Action (VLA) models have shown strong progress in end-to-end robot control by directly mapping visual observations and language instructions to actions~\cite{rt2,palme,openvla,atomvla,llava-vla,pi0,pi05}. Motivated by the strengths of both lines, recent unified world action models seek to jointly predict future observations and future actions within a single generative framework, offering a path toward general-purpose robot control that combines visual foresight with action generation~\cite{unified_wm,li2026causal,bi2025motusunifiedlatentaction,pertsch2025fast,gigaworld,dreamzero}.

Despite this promise, decoding reliable actions from such models still requires substantial robot-domain adaptation. We argue that a key limitation is structural rather than purely statistical. Predicting future RGB frames answers what the scene may look like, but control additionally depends on where the robot should interact and why that interaction is useful for the task. This type of spatially grounded reasoning has long been recognized as important in robotic manipulation~\cite{shridhar2022cliport,mo2021where2act,shridhar2023peract,wi2023calamari}. In cluttered manipulation, these signals are sparse, while future RGB representations are dense and dominated by appearance details that are often irrelevant to the next motor command. As a result, action decoding directly from future visual latents forces the model to recover manipulation intent implicitly from a representation that is not optimized for control.

To bridge this gap, we introduce \textbf{\method}, an intent-aware unified world action model that establishes an explicit spatial interface between future prediction and action generation. Rather than decoding actions directly from future visual representations, \method predicts an aligned action-based spatial value map that encodes the interaction structure required for task execution. This intermediate representation transforms future scene dynamics into a control-oriented spatial prior, providing a more direct pathway for action generation. Built on a shared pretrained video prior, \method jointly models future frames and future value maps, while an \emph{intent-causal attention} mechanism forces the action branch to access future information only through the predicted value map rather than directly through future visual tokens. In this way, the model separates scene evolution from manipulation intent without sacrificing their alignment.

We further develop a self-distillation RL post-training stage that freezes the video and value branches and optimizes only the action head using dense rewards induced by projected value-map responses together with sparse task-level rewards. To support training and evaluation, we build a large-scale simulation dataset with synchronized multi-view videos, action sequences, and value-map annotations. Experiments on the RoboTwin 2.0 benchmark~\cite{robotwin2.0} show that explicit spatial intent modeling consistently improves unified world action learning over strong baselines, while also yielding a more interpretable action-generation process through localized stage-wise interaction regions.

In summary, this work makes three main contributions:
\begin{itemize}[leftmargin=*]
    \item \textbf{Intent-aware unified world action model.}
    An explicit spatial value-map interface is introduced between future prediction and action decoding, enabling the model to capture task-relevant interaction structure in a control-oriented form.

    \item \textbf{A unified training framework for spatially grounded control.}
    Joint frame-value generation, intent-causal attention, and post-training self-distillation RL are integrated into a single framework that improves action learning without disturbing the prior training video.

    \item \textbf{Large-scale simulation dataset and strong benchmark performance.}
    We construct a large-scale simulation dataset of 30K manipulation trajectories with synchronized multi-view videos, actions, and value-map annotations. On the RoboTwin 2.0 benchmark~\cite{robotwin2.0}, AIM achieves \textbf{94.0\%} and \textbf{92.1\%} average success rates under the Easy and Hard settings, respectively, while consistently outperforming prior external baselines.
\end{itemize}

\section{Related Work}
\noindent\textbf{Video Generation Model for Robot Learning.}
Recent advances in video generation have made pretrained visual dynamics models attractive for robot learning, as they provide transferable priors over motion and long-horizon scene evolution that are difficult to obtain from robot-only data. Following this trend, recent methods increasingly build robot policies on top of pretrained video generators rather than learning world models from scratch~\cite{hu2024vpp,liang2025videogenerators,dreamzero}. Our work follows this paradigm and adopts Wan2.2-TI2V-5B~\cite{wan2025wan} as the video-generation backbone.

\noindent\textbf{Unified World Action Model.}
A growing line of work jointly models future observations and actions within a unified architecture~\cite{unified_wm,li2026causal,pertsch2025fast,gigaworld,dreamzero}. LingBot-VA~\cite{li2026causal} learns video prediction and action generation in a shared latent space, while GigaWorld-Policy~\cite{gigaworld} emphasizes action-centered world-action modeling for efficient control. Fast-WAM~\cite{pertsch2025fast} further shows that the benefit of world-action modeling may come more from video co-training than explicit future imagination at test time. Our method differs from this line by introducing an explicit spatial intermediate interface between future prediction and action decoding.

\noindent\textbf{Spatially Grounded Intermediate Representations.}
Spatially grounded representations are useful in robotic manipulation~\cite{3dgs} because they localize where interaction should occur. Prior work has studied actionable regions and voxel-aligned action spaces~\cite{mo2021where2act,shridhar2023peract}, while CLIPort~\cite{shridhar2022cliport} and CALAMARI~\cite{wi2023calamari} further demonstrate the importance of spatial grounding for object-centric and contact-rich manipulation. In contrast, we integrate spatial value prediction directly into a unified generative world-action model rather than using it as a standalone policy head or auxiliary perception module.

\begin{figure}[t]
  \centering
  \includegraphics[width=0.95\linewidth]{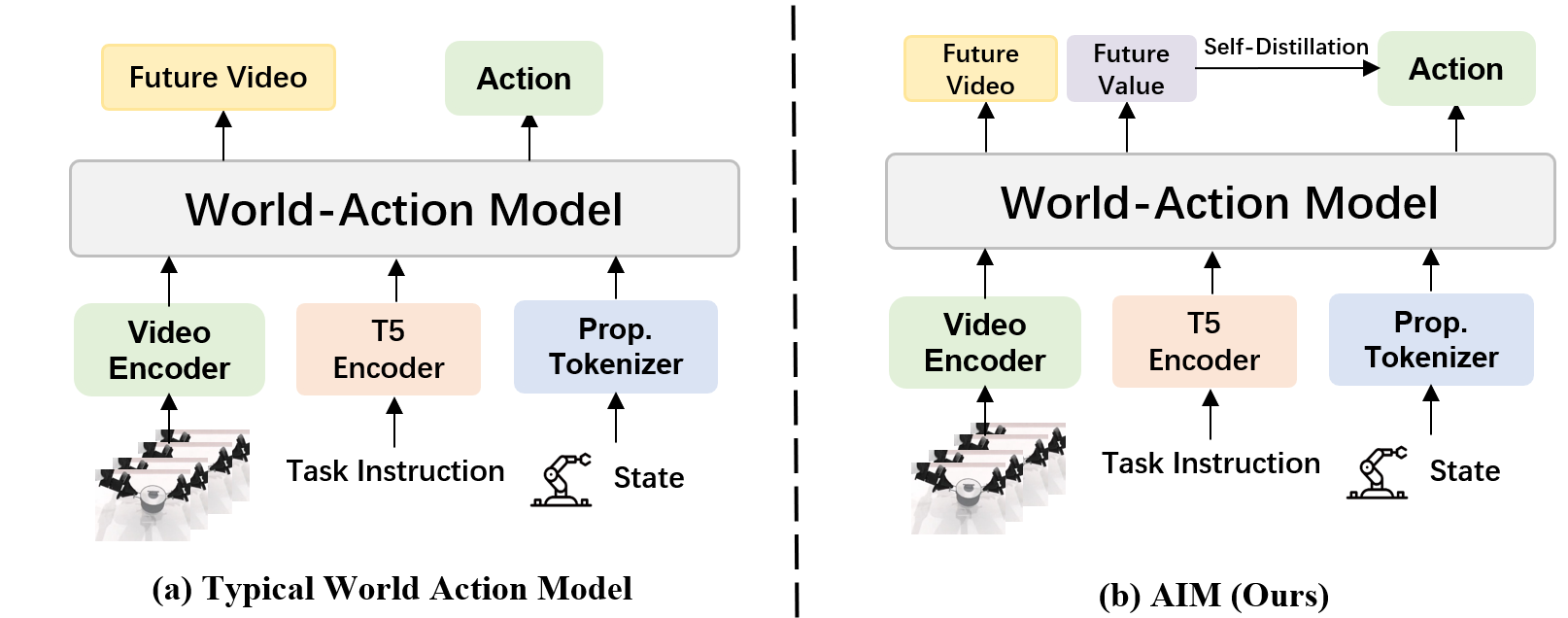}
  \caption{ (a) A typical unified world action model decodes actions directly from shared future visual representations. (b) \textbf{AIM (Ours)} introduces a spatial value-map interface for action decoding.}
  \label{fig:typical-vs-aim}
\end{figure}

\section{Overview}
\label{sec:overview}

\noindent\textbf{Problem Formulation.}
Traditional Vision-Language-Action (VLA) models~\cite{pi0,pi05,atomvla,llava-vla,openvla,rt2} learn action distributions directly from demonstrations. While effective for imitation, they do not explicitly model how robot actions shape future observations. In contrast, \method is a pretrained video-based world action model that imagines the near future by jointly modeling visual dynamics and action-relevant intent.
We consider manipulation trajectories $\tau=\{(o_t,a_t)\}_{t=1}^{T}$, where $o_t$ denotes synchronized multi-view observations and $a_t$ denotes robot actions. Given a history window $\mathcal{H}_t=\{o_{t-k:t}, a_{t-k:t-1}\}$, \method predicts a horizon-$h$ chunk of future RGB frames, spatial value maps, and robot actions, denoted by $X^+$, $M^+$, and $A^+$, respectively. Each future value map $m_t \in [0,1]^{H \times W \times 3}$ is spatially aligned with the corresponding RGB frame $x_t$ and highlights task-relevant interaction regions, serving as an explicit interface between future world modeling and action generation. We factorize the predictive distribution as
\begin{equation}
p(X^+, M^+, A^+ \mid \mathcal{H}_t)
=
p(X^+, M^+ \mid \mathcal{H}_t)\,
p(A^+ \mid \mathcal{H}_t, M^+).
\end{equation}
This factorization reflects the core design of \method: it first predicts future visual states together with aligned spatial value maps, and then decodes control actions conditioned on the predicted value maps rather than raw future RGB representations. The value map thus serves as the future-facing interface for action generation, bridging visual foresight and control. During deployment, \method performs autoregressive chunk-wise rollout with a transformer KV cache, appending each newly predicted world-action chunk to the prefix without recomputing the full history.

\noindent\textbf{Architecture Overview.}
\method consists of a video generation model, initialized from Wan2.2-TI2V-5B~\cite{wan2025wan}, and an action head with the same depth but a smaller hidden width for action denoising. Under the mixture-of-transformers design, the two branches share only the self-attention sublayer in each block, while maintaining separate feed-forward transformations.

At each rollout step, future RGB, value-map, and action tokens are initialized from noise. The video generation model jointly denoises RGB and value-map tokens along the same flow-matching trajectory, decoding them into future RGB observations and aligned spatial value maps, while the action head denoises action tokens into continuous control vectors. The three token streams interact through shared self-attention under a designed mask that allows future world tokens to access observation history while routing future information for action decoding through the value-map pathway, which we term \emph{intent-causal attention}. The predicted value map thus serves as an intent-aligned bridge between future world modeling and action generation.
\begin{figure}[t]
  \centering
  \includegraphics[width=0.95\linewidth]{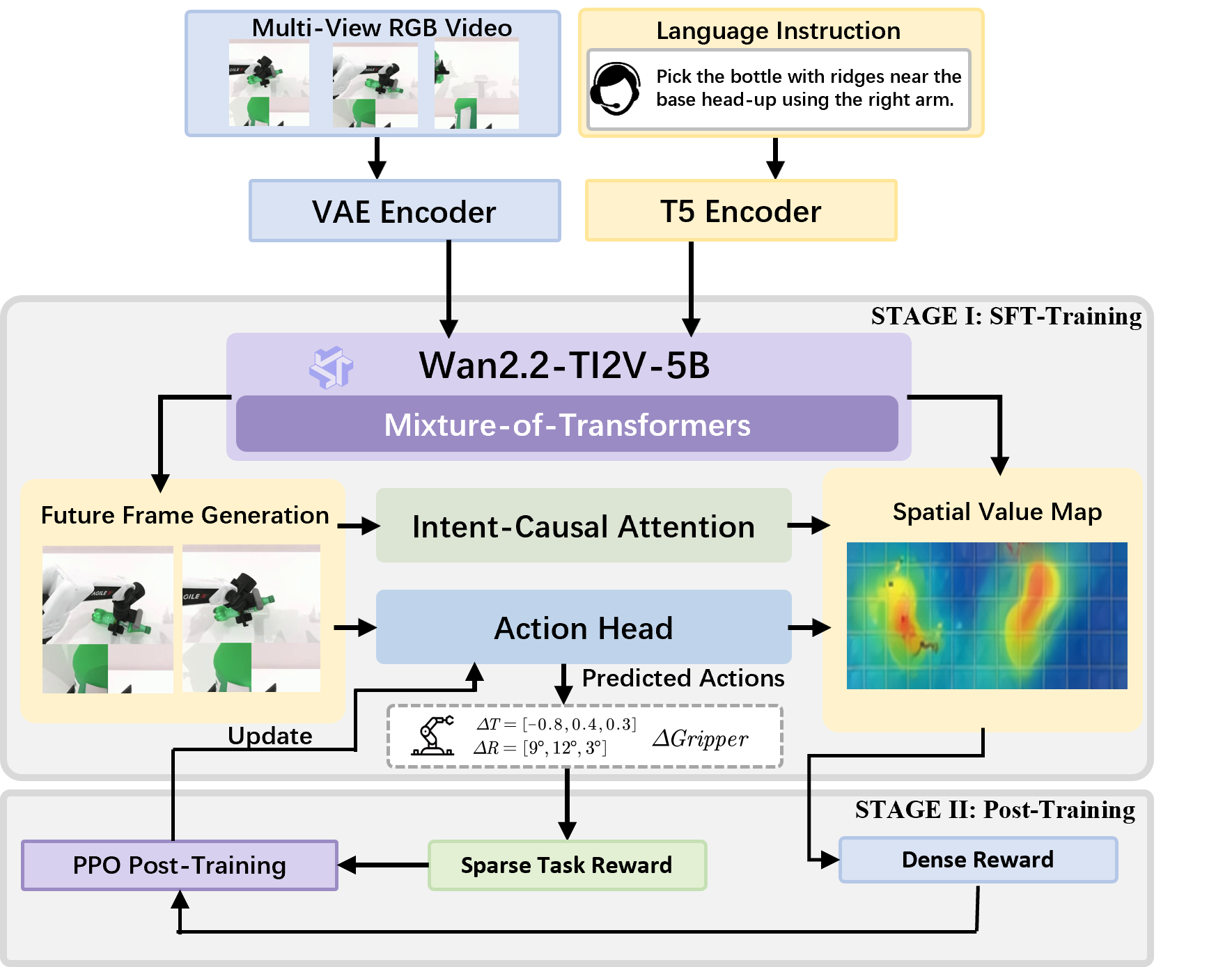}
  \caption{\textbf{Framework of AIM.} Given multi-view RGB videos and a language instruction, AIM jointly learns future frame generation, action prediction, and spatial value map estimation in Stage I, where intent-causal attention transfers task-relevant intention from visual prediction to action generation; in Stage II, the policy is further optimized with GRPO using both sparse and dense rewards.}
  \label{fig:typical-vs-aim}
\end{figure}

\section{Method}
\label{sec:method}
We contrast \method with a representative previous world action model as follows. Traditional WAMs either use a two-stage pipeline in which a video model first predicts the future state and then passes it to an action model, or use a one-stage design in which future state prediction and action prediction are conditioned on the same observation stream. In contrast, \method augments the video branch with an explicit value-map prediction pathway, so the action model does not need to infer inverse-dynamics cues directly from dense RGB futures.

\subsection{Model Architecture}
\noindent\textbf{Tokenization.} \method is built by adapting the pretrained video generation model Wan2.2-TI2V-5B~\cite{wan2025wan} to a world action model setting. To preserve the body of the pretrained video model without redesigning its visual input interface, we follow the multi-view packing strategy used in LingBot-VA~\cite{li2026causal} and merge the three onboard camera views into a single T-pose canvas: the head camera is placed at the top, and the left and right wrist cameras are placed on the corresponding sides. For action representation, we follow the recent trend of improving VLA action token efficiency through structured tokenization design~\cite{pertsch2025fast}. Let $\tilde{x}_t$ denote the packed RGB observation and $\tilde{m}_t$ denote the packed RGB ASVM at time $t$. We encode them with the pretrained Wan2.2 VAE encoder $E_{\mathrm{vae}}$ as
\begin{equation}
z_t^{o} = E_{\mathrm{vae}}(\tilde{x}_t), \qquad
z_t^{m} = E_{\mathrm{vae}}(\tilde{m}_t),
\end{equation}
where $z_t^{o}$ are observation tokens and $z_t^{m}$ are value-map tokens. Each ASVM is aligned with the corresponding future frame and uses the same T-pose layout as the RGB observation. At initialization, the value-map input is a pure black image, so the model starts from a null value prior and predicts the task-relevant spatial structure through denoising. Sharing the VAE between RGB and ASVM streams keeps value tokens geometrically aligned with visual tokens without modifying the pretrained tokenizer.

To align robot control with the visual token space, each dual-arm continuous action vector $a_t \in \mathbb{R}^{d_a}$ is projected by a lightweight MLP $E_a$ into an action token, while the text instruction $c$ is encoded by a pretrained T5 encoder~\cite{raffel2020t5} into language tokens:
\begin{equation}
z_t^{a} = E_a(a_t), \qquad z^{\ell} = E_{\mathrm{t5}}(c).
\end{equation}

These language features are injected only into the Video Model through cross-attention, so task semantics shape future world prediction without directly modifying the action branch. We denote the autoregressive prefix by
\begin{equation}
\mathcal{H}^{\mathrm{tok}}_t =
\left[z_{t-k:t}^{o},\, z_{t-k:t-1}^{a},\, z^{\ell}\right],
\end{equation}
which contains the current and recent observations, recent action history, and instruction tokens. This prefix provides both an estimate of the agent's physical state and the causal context used for future world action rollout.

\noindent\textbf{Architecture.} \method consists of a \emph{video generation model} for future RGB and value-map generation, and an \emph{action head} for future action generation. The video generation model is initialized from Wan2.2, while the action head uses the same depth but a smaller hidden width. At rollout step $t$, we initialize three future token groups from Gaussian noise,
\begin{equation}
\hat{z}_{0}^{x}, \hat{z}_{0}^{m}, \hat{z}_{0}^{a} \sim \mathcal{N}(0, I),
\end{equation}
where $\hat{z}_{0}^{x}$ denotes future RGB tokens, $\hat{z}_{0}^{m}$ denotes future value-map tokens, and $\hat{z}_{0}^{a}$ denotes future action tokens. The value stream additionally receives a learned value noise token $n^{m}$, so the actual input to the value branch is $[\hat{z}_{0}^{m}, n^m]$. Future RGB tokens and future value-map tokens are jointly denoised by the video generation model along the same flow-matching trajectory~\cite{lipman2023flowmatching}, while future action tokens are denoised by the action head. Their decoded outputs are
\begin{equation}
\hat{X}^{+} = D_x(z^{x}), \qquad
\hat{M}^{+} = D_m(z^{m}), \qquad
\hat{A}^{+} = D_a(z^{a}),
\end{equation}
where $\hat{X}^{+}$, $\hat{M}^{+}$, and $\hat{A}^{+}$ are the predicted future RGB frames, spatial value maps, and continuous dual-arm actions, respectively.

We adopt a \mot architecture so that action tokens can be tightly coupled with visual tokens while preserving the computational structure of the pretrained backbone. Let $h_s^\ell$ denote the hidden states of stream $s \in \{x,m,a\}$ at transformer layer $\ell$. Each stream computes its own projections
\begin{equation}
Q_s^\ell = h_s^\ell W_{Q,s}^\ell,\qquad
K_s^\ell = h_s^\ell W_{K,s}^\ell,\qquad
V_s^\ell = h_s^\ell W_{V,s}^\ell,
\end{equation}
which preserve stream-specific feature spaces. These stream-specific queries, keys, and values are then projected to a common attention dimension and participate in a shared masked self-attention operation. The resulting attended features are projected back to each stream and added through residual connections, while the feed-forward transformations remain branch-specific. Language features from T5 are injected only into the video generation model through cross-attention,
\begin{equation}
h_x^\ell \leftarrow h_x^\ell + \mathrm{CA}(h_x^\ell, z^{\ell}),
\end{equation}
where $\mathrm{CA}(\cdot,\cdot)$ denotes cross-attention with the instruction tokens. The action head receives task semantics only through the shared world and value representations rather than through direct language conditioning. The overall objective is a weighted sum of the RGB flow-matching loss, the value-map flow-matching loss, and the inverse-dynamics loss for the action head:
\begin{equation}
\mathcal{L} = \mathcal{L}_{\mathrm{rgb}} + \lambda_m \mathcal{L}_{\mathrm{map}} + \lambda_a \mathcal{L}_{\mathrm{act}}.
\end{equation}
Here $\mathcal{L}_{\mathrm{rgb}}$ and $\mathcal{L}_{\mathrm{map}}$ supervise the flow-matching velocity fields for future RGB and value-map denoising, while $\mathcal{L}_{\mathrm{act}}$ supervises inverse-dynamics action prediction from the action head. At inference time, \method operates autoregressively and naturally supports KV caching. Only newly appended real observations and newly predicted tokens require fresh attention computation, while all earlier history tokens are reused from the cache, substantially improving long-horizon rollout efficiency.

\subsection{Intent-Causal Self-Attention}
Directly training a generic \mot world action network makes inverse-dynamics learning difficult, because the action head must extract sparse action-relevant cues from dense future RGB appearance generated by the video generation model, especially in cluttered scenes. We address this by jointly modeling future states and their associated value distributions: the video generation model predicts both future RGB frames and future ASVMs, whose value maps provide a more compact and action-relevant representation than raw RGB appearance, making inverse-dynamics learning substantially simpler and more explicit.

We implement this idea with \emph{intent-causal self-attention}, a designed visibility mask over the shared self-attention layers. Let $\mathcal{V}_x$, $\mathcal{V}_m$, and $\mathcal{V}_a$ denote the visible token sets for future video, value, and action streams, respectively:
\begin{equation}
\begin{aligned}
\mathcal{V}_x &= \left[z_t^{o},\, z_{t-k:t-1}^{o},\, z_{t-k:t-1}^{a},\, z^{\ell},\, z^{x}\right],\\
\mathcal{V}_m &= \left[z_t^{o},\, z_{t-k:t-1}^{o},\, z^{x},\, z^{m}\right],\\
\mathcal{V}_a &= \left[z_t^{o},\, z_{t-k:t-1}^{a},\, z^{m},\, z^{a}\right].
\end{aligned}
\end{equation}
Here $z^{x}$, $z^{m}$, and $z^{a}$ denote the future video, value, and action tokens being denoised at the current rollout step. The shared masked attention update for stream $s$ is
\begin{equation}
\tilde{h}_s^\ell
=
\mathrm{Attn}\!\bigl(
Q_s^\ell,\,
K(\mathcal{V}_s),\,
V(\mathcal{V}_s)
\bigr),
\end{equation}
followed by the stream-specific residual projection back to the original hidden space.

This mask has a clear semantic effect. Future video tokens can attend to the current observation, instruction tokens, and past observation and action tokens, so the video generation model predicts task-conditioned future world states. Future value tokens can attend to the current observation, past observations, and future video tokens, ensuring that value prediction remains anchored to the sampled future state. Action tokens can attend to historical action tokens, the current observation, and future value tokens, but not future RGB tokens directly. As a result, task semantics enter the video generation model through T5-conditioned cross-attention, future state information is consolidated into the value stream, and the action head accesses future information only through the predicted value representation.

\subsection{Self-Distillation RL Post-Training}

Supervised training teaches the action head to imitate the action distribution in the dataset, but it does not fully optimize control success under closed-loop execution. We therefore introduce a reinforcement-learning post-training stage that updates only the action head while keeping the video generator and value-map head frozen. This selective finetuning stabilizes the learned visual world model and avoids catastrophic drift in future frame and value-map prediction.

Let $\pi_\phi(a_t \mid \mathcal{H}_t, m_{t+1:t+h})$ denote the action head after pretraining. During RL post-training, the model samples actions in the environment and receives two reward components:
\begin{equation}
\begin{aligned}
r_t &= \lambda_d r_t^{\mathrm{dense}} + \lambda_s r_t^{\mathrm{sparse}},\\
r_t^{\mathrm{dense}} &= M_t(\Pi(p_t)),
\end{aligned}
\end{equation}
where $r_t^{\mathrm{sparse}}$ is the environment-level task success or completion signal, $p_t$ is the predicted action landing point or end-effector target, $\Pi(\cdot)$ is the camera projection function, and $M_t$ is the predicted value map at time step $t$. Intuitively, the action head is rewarded when the projected action falls on high-value interaction regions predicted by the frozen value head.

We optimize only the action head with GRPO \cite{shao2024deepseekmath}, of which the objective is
\begin{equation}
\begin{aligned}
\mathcal{L}_{\mathrm{GRPO}}(\phi)
&=
\mathbb{E}_t \left[
\min\!\left(
\rho_t(\phi)\hat{A}_t,\,
\mathrm{clip}\!\left(\rho_t(\phi), 1-\epsilon, 1+\epsilon\right)\hat{A}_t
\right)
\right],\\
\rho_t(\phi)
&=
\frac{\pi_\phi(a_t \mid \mathcal{H}_t, m_{t+1:t+h})}
{\pi_{\phi_{\mathrm{old}}}(a_t \mid \mathcal{H}_t, m_{t+1:t+h})},
\end{aligned}
\end{equation}
where $\hat{A}_t$ is the advantage estimated from returns under the combined reward $r_t$, and $\epsilon$ is the clipping coefficient. Because the dense reward is derived from the model's own predicted spatial values, this stage can be viewed as a form of self-distillation: the pretrained value head supervises the action head under online interaction without requiring additional human labels.

\begin{table*}[htbp]
\centering
\captionsetup{font=small, labelfont=bf, skip=4pt}
\caption{Per-task success rates~\textbf{SR} on 50 RoboTwin simulation tasks under \textbf{Easy} and ~\textbf{Hard} settings.}
\label{tab:stage1-stage2-baselines-final}

\small
\setlength{\tabcolsep}{3.8pt}
\renewcommand{\arraystretch}{1.06}

\begin{tabular}{>{\itshape\raggedright\arraybackslash}p{3.35cm}cccccccccc}
\toprule
\multirow{2}{*}{Simulation Task}
& \multicolumn{2}{c}{$\pi_{0.5}$}
& \multicolumn{2}{c}{X-VLA}
& \multicolumn{2}{c}{Motus}
& \multicolumn{2}{c}{Stage1}
& \multicolumn{2}{c}{\textbf{AIM (Ours)}} \\
\cmidrule(lr){2-3}\cmidrule(lr){4-5}\cmidrule(lr){6-7}\cmidrule(lr){8-9}\cmidrule(lr){10-11}
& Easy & Hard
& Easy & Hard
& Easy & Hard
& Easy & Hard
& Easy & Hard \\
\midrule
Adjust Bottle & \bestcell{100\%} & 99\% & \bestcell{100\%} & 99\% & 89\% & 93\% & 98\% & 99\% & \bestcell{100\%} & \bestcell{100\%} \\
Beat Block Hammer & 96\% & 93\% & 92\% & 88\% & 95\% & 88\% & 98\% & \bestcell{100\%} & \bestcell{100\%} & \bestcell{100\%} \\
Blocks Ranking RGB & 92\% & 85\% & 83\% & 83\% & \bestcell{99\%} & 97\% & 91\% & 77\% & 92\% & 77\% \\
Blocks Ranking Size & 49\% & 26\% & 67\% & 74\% & \bestcell{75\%} & 63\% & 47\% & 44\% & 47\% & 43\% \\
Click Alarmclock & 98\% & 89\% & 99\% & 99\% & \bestcell{100\%} & \bestcell{100\%} & 98\% & 99\% & \bestcell{100\%} & \bestcell{100\%} \\
Click Bell & 99\% & 66\% & \bestcell{100\%} & \bestcell{100\%} & \bestcell{100\%} & \bestcell{100\%} & 98\% & 99\% & \bestcell{100\%} & \bestcell{100\%} \\
Dump Bin Bigbin & 92\% & 97\% & 79\% & 77\% & 95\% & 91\% & 98\% & \bestcell{100\%} & \bestcell{100\%} & \bestcell{100\%} \\
Grab Roller & \bestcell{100\%} & \bestcell{100\%} & \bestcell{100\%} & \bestcell{100\%} & \bestcell{100\%} & \bestcell{100\%} & 98\% & 99\% & \bestcell{100\%} & \bestcell{100\%} \\
Handover Block & 66\% & 57\% & 73\% & 37\% & 86\% & 73\% & 92\% & 89\% & \bestcell{93\%} & 90\% \\
Handover Mic & \bestcell{98\%} & 97\% & 0\% & 0\% & 78\% & 63\% & 82\% & 82\% & 83\% & 81\% \\
Hanging Mug & 18\% & 17\% & 23\% & 27\% & 38\% & 38\% & \bestcell{43\%} & \bestcell{43\%} & \bestcell{43\%} & 42\% \\
Lift Pot & 96\% & 85\% & 99\% & \bestcell{100\%} & 96\% & 99\% & 98\% & \bestcell{100\%} & \bestcell{100\%} & \bestcell{100\%} \\
Move Can Pot & 51\% & 55\% & 89\% & 86\% & 34\% & 74\% & 99\% & 97\% & \bestcell{100\%} & 98\% \\
Move Pillbottle Pad & 84\% & 61\% & 73\% & 71\% & 93\% & 96\% & 97\% & \bestcell{99\%} & 97\% & 98\% \\
Move Playingcard Away & 96\% & 84\% & 93\% & 98\% & \bestcell{100\%} & 96\% & 98\% & \bestcell{100\%} & \bestcell{100\%} & \bestcell{100\%} \\
Move Stapler Pad & 56\% & 42\% & 78\% & 73\% & 83\% & 85\% & 91\% & 83\% & \bestcell{92\%} & 84\% \\
Open Laptop & 90\% & 96\% & 93\% & \bestcell{100\%} & 95\% & 91\% & 98\% & \bestcell{100\%} & \bestcell{100\%} & \bestcell{100\%} \\
Open Microwave & 34\% & 77\% & 79\% & 71\% & \bestcell{95\%} & 91\% & 83\% & 80\% & 83\% & 79\% \\
Pick Diverse Bottles & 81\% & 71\% & 58\% & 36\% & 90\% & 91\% & 99\% & 97\% & \bestcell{100\%} & 98\% \\
Pick Dual Bottles & 93\% & 63\% & 47\% & 36\% & \bestcell{96\%} & 90\% & 92\% & 90\% & 93\% & 91\% \\
Place A2B Left & 87\% & 82\% & 48\% & 49\% & 82\% & 79\% & 93\% & 91\% & \bestcell{94\%} & 92\% \\
Place A2B Right & 87\% & 84\% & 36\% & 36\% & \bestcell{90\%} & 87\% & 89\% & 89\% & \bestcell{90\%} & 88\% \\
Place Bread Basket & 77\% & 64\% & 81\% & 71\% & 91\% & \bestcell{94\%} & 92\% & 90\% & 93\% & 91\% \\
Place Bread Skillet & 85\% & 66\% & 77\% & 67\% & 86\% & 83\% & 98\% & \bestcell{100\%} & \bestcell{100\%} & \bestcell{100\%} \\
Place Burger Fries & 94\% & 87\% & 94\% & 94\% & 98\% & 98\% & 98\% & \bestcell{100\%} & \bestcell{100\%} & \bestcell{100\%} \\
Place Can Basket & 62\% & 62\% & 49\% & 52\% & \bestcell{81\%} & 76\% & 78\% & 77\% & 78\% & 76\% \\
Place Cans Plasticbox & 94\% & 84\% & 97\% & 98\% & 98\% & 94\% & 98\% & \bestcell{100\%} & \bestcell{100\%} & \bestcell{100\%} \\
Place Container Plate & 99\% & 95\% & 97\% & 95\% & 98\% & 99\% & \bestcell{100\%} & 96\% & \bestcell{100\%} & 97\% \\
Place Dual Shoes & 75\% & 75\% & 79\% & 88\% & 93\% & 87\% & \bestcell{100\%} & 99\% & \bestcell{100\%} & 98\% \\
Place Empty Cup & \bestcell{100\%} & 99\% & \bestcell{100\%} & 98\% & 99\% & 98\% & 98\% & \bestcell{100\%} & \bestcell{100\%} & \bestcell{100\%} \\
Place Fan & 87\% & 85\% & 80\% & 75\% & 91\% & 87\% & \bestcell{93\%} & 89\% & \bestcell{93\%} & 90\% \\
Place Mouse Pad & 60\% & 39\% & 70\% & 70\% & 66\% & 68\% & \bestcell{97\%} & 96\% & \bestcell{97\%} & 95\% \\
Place Object Basket & 80\% & 76\% & 44\% & 39\% & 81\% & 87\% & \bestcell{93\%} & 88\% & \bestcell{93\%} & 89\% \\
Place Object Scale & 86\% & 80\% & 52\% & 74\% & 88\% & 85\% & \bestcell{100\%} & 97\% & \bestcell{100\%} & 98\% \\
Place Object Stand & 91\% & 85\% & 86\% & 88\% & 98\% & 97\% & 98\% & \bestcell{100\%} & \bestcell{100\%} & \bestcell{100\%} \\
Place Phone Stand & 81\% & 81\% & \bestcell{88\%} & 87\% & 87\% & 86\% & 82\% & 81\% & 82\% & 80\% \\
Place Shoe & 92\% & 93\% & 96\% & 95\% & 99\% & 97\% & 98\% & \bestcell{100\%} & \bestcell{100\%} & \bestcell{100\%} \\
Press Stapler & 87\% & 83\% & 92\% & \bestcell{98\%} & 93\% & \bestcell{98\%} & 96\% & 95\% & 96\% & 94\% \\
Put Bottles Dustbin & \bestcell{84\%} & 79\% & 74\% & 77\% & 81\% & 79\% & 80\% & 75\% & 80\% & 74\% \\
Put Object Cabinet & 80\% & 79\% & 46\% & 48\% & \bestcell{88\%} & 71\% & 81\% & 75\% & 81\% & 74\% \\
Rotate QRcode & 89\% & 87\% & 34\% & 33\% & 89\% & 73\% & 98\% & 99\% & \bestcell{100\%} & 98\% \\
Scan Object & 72\% & 65\% & 14\% & 36\% & 67\% & 66\% & 98\% & 97\% & \bestcell{100\%} & 98\% \\
Shake Bottle Horizontally & 99\% & 99\% & \bestcell{100\%} & \bestcell{100\%} & \bestcell{100\%} & 98\% & 98\% & \bestcell{100\%} & \bestcell{100\%} & \bestcell{100\%} \\
Shake Bottle & 99\% & 97\% & 99\% & \bestcell{100\%} & \bestcell{100\%} & 97\% & 98\% & \bestcell{100\%} & \bestcell{100\%} & \bestcell{100\%} \\
Stack Blocks Three & 91\% & 76\% & 6\% & 10\% & 91\% & 95\% & \bestcell{100\%} & 99\% & \bestcell{100\%} & 98\% \\
Stack Blocks Two & 97\% & \bestcell{100\%} & 92\% & 87\% & \bestcell{100\%} & 98\% & 98\% & \bestcell{100\%} & \bestcell{100\%} & \bestcell{100\%} \\
Stack Bowls Three & 77\% & 71\% & 76\% & 86\% & 79\% & 87\% & \bestcell{100\%} & 99\% & \bestcell{100\%} & 98\% \\
Stack Bowls Two & 95\% & 96\% & 96\% & 93\% & 98\% & 98\% & \bestcell{100\%} & 97\% & \bestcell{100\%} & 98\% \\
Stamp Seal & 79\% & 55\% & 76\% & 82\% & 93\% & 92\% & \bestcell{100\%} & \bestcell{100\%} & \bestcell{100\%} & \bestcell{100\%} \\
Turn Switch & 62\% & 54\% & 40\% & 61\% & 84\% & 78\% & \bestcell{100\%} & 99\% & \bestcell{100\%} & 98\% \\

\midrule
\multicolumn{1}{l}{\textbf{\upshape Average}}
& 82.74\% & 76.76\%
& 72.80\% & 72.84\%
& 88.66\% & 87.02\%
& 93.0\% & 92.0\%
& \bestcell{94.0\%} & \bestcell{92.1\%} \\
\bottomrule
\end{tabular}
\end{table*}

\section{Dataset and Value-Map Annotation}
\label{sec:dataset}

\noindent\textbf{Large-Scale Simulation Dataset.}
We build a large-scale simulation dataset of 30K manipulation trajectories collected on RoboTwin 2.0 Benchmark~\cite{robotwin2.0}. Each trajectory contains synchronized multi-view videos, action sequences, task identifiers, and per-step value-map annotations.

A simulation-only setting is well-suited to our study for two reasons. First, it provides the scale needed for training unified world-action models while supporting automatic label generation from contact events and physics states. Second, it offers precise geometric information and controllable scene diversity, enabling consistent value-map annotation across tasks, viewpoints, and object configurations. This setting allows us to study spatially grounded world-action modeling under controlled conditions at scale.

\noindent\textbf{Pick Value-Map Annotation.}
For pick tasks, we record the contact surface point cloud when the gripper establishes effective grasp contact with the target object. The simulator's contact-detection API provides a set of contact vertices on the interacting surfaces. We project these points into the image plane using the calibrated camera projection matrix and then apply Gaussian smoothing to obtain a continuous-value heat map. This annotation marks the \emph{grasp affordance region}: the spatial area where the object and end-effector engage in successful physical interaction. The Gaussian kernel width is adjusted dynamically according to camera parameters and the distance between the projected point and the camera, so that the label maintains a meaningful image-space support size across different viewpoints and depths.

\noindent\textbf{Place Value-Map Annotation.}
For place tasks, we detect placement completion when the manipulated object reaches a stable configuration, defined by a small center-of-mass velocity threshold. We then extract the contact region between the grasped object and the target support surface, project the contact area into the image plane, and generate a placement feasibility heat map. This map marks the \emph{placement contact region}: the spatial location where the held object should make contact with the environment to satisfy the placement goal.

\begin{table}[t]
\centering
\caption{RoboTwin 2.0 average success rate~\textbf{(SR)} under \textbf{Easy} and \textbf{Hard} settings. }
\label{tab:robotwin_avg_clean_rand_methods_as_columns}
\small
\setlength{\tabcolsep}{4pt}
\begin{tabular}{lccccccccc}
\toprule
Setting & $\pi_{0}$ & $\pi_{0.5}$ & X-VLA & Motus & Fast-WAM & Giga-World & LingBot-VA & Stage1 & ~\textbf{AIM (Ours)} \\
\midrule
Easy & 65.9\% & 82.7\% & 72.8\% & 88.7\% & 91.9\% & 87.0\% & 92.9\% & 93.0\% & \textbf{94.0\%} \\
Hard & 58.4\% & 76.8\% & 72.8\% & 87.0\% & 91.8\% & 85.0\% & 91.6\% & 92.0\% & \textbf{92.1\%} \\
\midrule
Average & 62.2\% & 79.8\% & 72.8\% & 87.8\% & 91.8\% & 86.0\% & 92.2\% & 92.5\% & \textbf{93.1\%} \\
\bottomrule
\end{tabular}
\end{table}

\begin{figure}[t]
  \centering
  \includegraphics[width=0.99\linewidth]{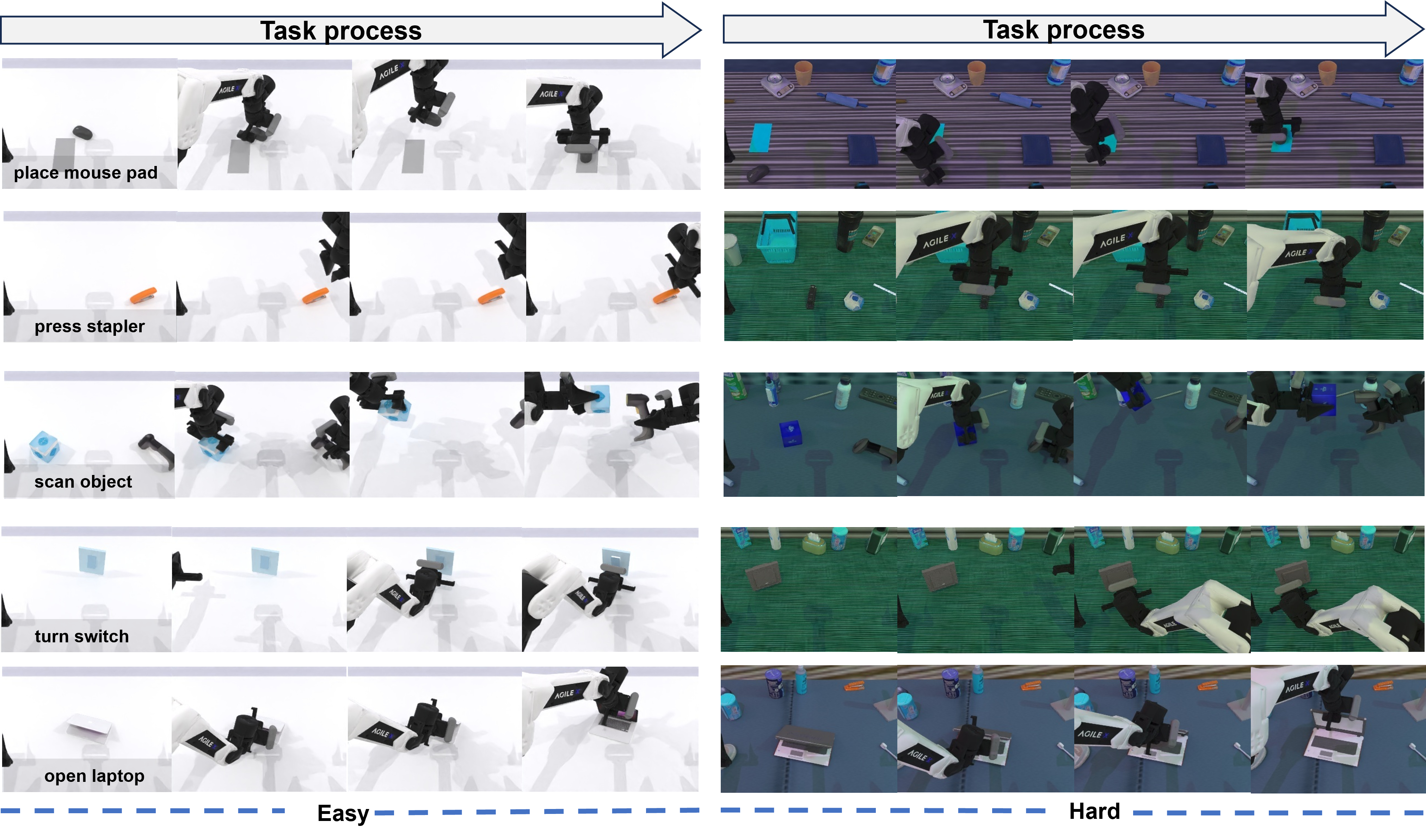}
  \caption{Representative task execution processes in Robotwin 2.0 benchmark~\cite{robotwin2.0}, including \textit{place mouse pad}, \textit{press stapler}, \textit{scan object}, \textit{turn switch}, and \textit{open laptop}. The left column shows examples under \textbf{Easy} setting, while the right column shows examples under \textbf{Hard} setting.}
  \label{fig:typical-vs-aim}
\end{figure}

\section{Experiments}

\noindent\textbf{Experimental Setup.}
We evaluate \method on 50 RoboTwin simulation tasks under Easy and Hard settings. Unless otherwise noted, all models are trained on the same 30K RoboTwin simulation dataset and evaluated under identical task definitions and success criteria. The video generation model is initialized from Wan2.2-TI2V-5B~\cite{wan2025wan}.

\noindent\textbf{Baselines and Metrics.}
We select $\pi_{0.5}$~\cite{pi05}, X-VLA~\cite{xVLA}, and Motus~\cite{bi2025motusunifiedlatentaction} as our baselines, and additionally report Stage1, our supervised model before RL post-training, to isolate the effect of self-distillation. We use success rate (\textbf{SR}) as the primary metric. During RL post-training, the action head is initialized from the Stage1 checkpoint, while the video generation model and value-map head remain frozen.

\noindent\textbf{Main Results and Analysis.}
As shown in Table~\ref{tab:stage1-stage2-baselines-final}, \method achieves average SRs of \textbf{94.0\%} and \textbf{92.1\%} under the Easy and Hard settings, respectively, outperforming all external baselines~\cite{xVLA,pi0,pi05,bi2025motusunifiedlatentaction,yuan2026fastwam,gigaworld,li2026causal}. In particular, compared with Motus~\cite{bi2025motusunifiedlatentaction}, \method improves the average SR by \textbf{+5.3\%} and \textbf{+5.0\%} under Easy and Hard setting, respectively; compared with $\pi_{0.5}$~\cite{pi05}, the gains increase to \textbf{+11.3\%} and \textbf{+15.3\%}. Stage 1 reaches \textbf{93.0\%}/\textbf{92.0\%}, indicating that RL post-training brings further improvement. The largest gains in Table~\ref{tab:stage1-stage2-baselines-final} appear on contact-sensitive and stage-dependent tasks, such as \emph{Place Mouse Pad} (\textbf{97\%}/\textbf{95\%}), \emph{Scan Object} (\textbf{100\%}/\textbf{98\%}), and \emph{Turn Switch} (\textbf{100\%}/\textbf{98\%}), where accurate localization of task-relevant interaction regions is critical. This trend is also supported qualitatively: future frame predictions remain temporally aligned with the manipulation stage, value maps concentrate on meaningful interaction regions rather than generic saliency, and projected action targets fall within the corresponding high-value areas, suggesting that the gains indeed come from the intended spatial bridge rather than shortcut correlations.

\section{Conclusion}
This paper presented \method, an intent-aware unified world-action model that bridges future video generation and robot action decoding with explicit spatial value maps. Built on a pretrained video generator and a \mot architecture, \method jointly predicts future frames, future value maps, and future actions, while intent-causal attention ensures that future information reaches the action head only through the intent-aligned value representation. We further introduced a self-distillation RL post-training strategy that freezes the video generation model and value branch and updates only the action head using dense rewards derived from projected value-map responses.

Together with a 30K simulation dataset and evaluation on 50 RoboTwin 2.0~\cite{robotwin2.0} simulation tasks, these designs provide a coherent framework for studying spatially grounded world-action modeling. AIM achieves strong performance on the RoboTwin benchmark and consistently outperforms prior external baselines, showing that explicit spatial value modeling is a practical and effective bridge between RGB world modeling and robot action generation.
\clearpage
\bibliographystyle{plainnat}
\bibliography{Reference}

@article{li2026causal,
  title={Causal World Modeling for Robot Control},
  author={Li, Lin and Zhang, Qihang and Luo, Yiming and Yang, Shuai and Wang, Ruilin and Han, Fei and Yu, Mingrui and Gao, Zelin and Xue, Nan and Zhu, Xing and Shen, Yujun and Xu, Yinghao},
  journal={arXiv preprint arXiv:2601.21998},
  year={2026}
}

@article{gigaworld,
  title={GigaWorld-Policy: An Efficient Action-Centered World--Action Model},
  author={Ye, Angen and Wang, Boyuan and Ni, Chaojun and Huang, Guan and Zhao, Guosheng and Li, Hao and Li, Hengtao and Li, Jie and Lv, Jindi and Liu, Jingyu and Cao, Min and Li, Peng},
  journal={arXiv preprint arXiv:2603.17240},
  year={2026}
}

@article{wan2025wan,
  title={Wan: Open and Advanced Large-Scale Video Generative Models},
  author={Wan, Team and Wang, Ang and Ai, Baole and Wen, Bin and Mao, Chaojie and Xie, Chen-Wei and Chen, Di and Yu, Feiwu},
  journal={arXiv preprint arXiv:2503.20314},
  year={2025}
}

@article{raffel2020t5,
  title={Exploring the Limits of Transfer Learning with a Unified Text-to-Text Transformer},
  author={Raffel, Colin and Shazeer, Noam and Roberts, Adam and Lee, Katherine and Narang, Sharan and Matena, Michael and Zhou, Yanqi and Li, Wei and Liu, Peter J.},
  journal={Journal of Machine Learning Research},
  volume={21},
  number={140},
  pages={1--67},
  year={2020}
}

@article{lipman2023flowmatching,
  title={Flow Matching for Generative Modeling},
  author={Lipman, Yaron and Chen, Ricky T. Q. and Ben-Hamu, Heli and Nickel, Maximilian and Le, Matt},
  journal={arXiv preprint arXiv:2210.02747},
  year={2023}
}

@article{yuan2026fastwam,
  title={Fast-WAM: Do World Action Models Need Test-time Future Imagination?},
  author={Yuan, Tianyuan and Dong, Zibin and Liu, Yicheng and Zhao, Hang},
  journal={arXiv preprint arXiv:2603.16666},
  year={2026}
}

@article{shao2024deepseekmath,
  title={Deepseekmath: Pushing the limits of mathematical reasoning in open language models},
  author={Shao, Zhihong and Wang, Peiyi and Zhu, Qihao and Xu, Runxin and Song, Junxiao and Bi, Xiao and Zhang, Haowei and Zhang, Mingchuan and Li, YK and Wu, Yang and others},
  journal={arXiv preprint arXiv:2402.03300},
  year={2024}
}

@misc{bi2025motusunifiedlatentaction,
  title={Motus: A Unified Latent Action World Model},
  author={Bi, Hongzhe and Tan, Hengkai and Xie, Shenghao and Wang, Zeyuan and Huang, Shuhe and Liu, Haitian and Zhao, Ruowen and Zhu, Jun and others},
  year={2025},
  eprint={2512.13030},
  archivePrefix={arXiv},
  primaryClass={cs.CV},
  url={https://arxiv.org/abs/2512.13030}
}

@article{pertsch2025fast,
  title={Fast: Efficient action tokenization for vision-language-action models},
  author={Pertsch, Karl and Stachowicz, Kyle and Ichter, Brian and Driess, Danny and Nair, Suraj and Vuong, Quan and Mees, Oier and Finn, Chelsea and Levine, Sergey},
  journal={arXiv preprint arXiv:2501.09747},
  year={2025}
}

@article{liang2025videogenerators,
  title={Video Generators are Robot Policies}, 
  author={Liang, Junbang and Tokmakov, Pavel and Liu, Ruoshi and Sudhakar, Sruthi and Shah, Paarth and Ambrus, Rares and Vondrick, Carl},
  journal={arXiv preprint arXiv:2508.00795},
  year={2025}
}

@inproceedings{shridhar2022cliport,
  title     = {CLIPort: What and Where Pathways for Robotic Manipulation},
  author    = {Shridhar, Mohit and Manuelli, Lucas and Fox, Dieter},
  booktitle = {Proceedings of the 5th Conference on Robot Learning},
  series    = {Proceedings of Machine Learning Research},
  volume    = {164},
  pages     = {894--906},
  year      = {2022}
}

@inproceedings{wi2023calamari,
  title     = {CALAMARI: Contact-Aware and Language conditioned spatial Action MApping for contact-RIch manipulation},
  author    = {Wi, Youngsun and Merwe, Mark Van der and Florence, Pete and Zeng, Andy and Fazeli, Nima},
  booktitle = {Proceedings of The 7th Conference on Robot Learning},
  series    = {Proceedings of Machine Learning Research},
  volume    = {229},
  pages     = {2753--2771},
  year      = {2023}
}

@article{atomvla,
  title={AtomVLA: Scalable Post-Training for Robotic Manipulation via Predictive Latent World Models},
  author={Sun, Xiaoquan and Xu, Zetian and Cao, Chen and Liu, Zonghe and Sun, Yihan and Pang, Jingrui and Zhang, Ruijian and Yang, Zhen and Pang, Kang and He, Dingxin and others},
  journal={arXiv preprint arXiv:2603.08519},
  year={2026}
}

@article{robotwin2.0,
  title={Robotwin 2.0: A scalable data generator and benchmark with strong domain randomization for robust bimanual robotic manipulation},
  author={Chen, Tianxing and Chen, Zanxin and Chen, Baijun and Cai, Zijian and Liu, Yibin and Li, Zixuan and Liang, Qiwei and Lin, Xianliang and Ge, Yiheng and Gu, Zhenyu and others},
  journal={arXiv preprint arXiv:2506.18088},
  year={2025}
}

@article{dreamzero,
  title={World action models are zero-shot policies},
  author={Ye, Seonghyeon and Ge, Yunhao and Zheng, Kaiyuan and Gao, Shenyuan and Yu, Sihyun and Kurian, George and Indupuru, Suneel and Tan, You Liang and Zhu, Chuning and Xiang, Jiannan and others},
  journal={arXiv preprint arXiv:2602.15922},
  year={2026}
}

@article{llava-vla,
  title={Rethinking the Practicality of Vision-language-action Model: A Comprehensive Benchmark and An Improved Baseline}, 
  author={Wenxuan Song and Jiayi Chen and Xiaoquan Sun and Huashuo Lei and Yikai Qin and Wei Zhao and Pengxiang Ding and Han Zhao and Tongxin Wang and Pengxu Hou and Zhide Zhong and Haodong Yan and Donglin Wang and Jun Ma and Haoang Li},
  journal={arXiv preprint arXiv:2602.22663}, 
  year={2026},
}

@article{unified_wm,
  title={Unified world models: Coupling video and action diffusion for pretraining on large robotic datasets},
  author={Zhu, Chuning and Yu, Raymond and Feng, Siyuan and Burchfiel, Benjamin and Shah, Paarth and Gupta, Abhishek},
  journal={arXiv preprint arXiv:2504.02792},
  year={2025}
}

@inproceedings{mo2021where2act,
  title={Where2Act: From Pixels to Actions for Articulated 3D Objects},
  author={Mo, Kaichun and Guibas, Leonidas J. and Mukadam, Mustafa and Gupta, Abhinav and Tulsiani, Shubham},
  booktitle={Proceedings of the IEEE/CVF International Conference on Computer Vision (ICCV)},
  pages={6813--6823},
  year={2021}
}

@inproceedings{shridhar2023peract,
  title={Perceiver-Actor: A Multi-Task Transformer for Robotic Manipulation},
  author={Shridhar, Mohit and Manuelli, Lucas and Fox, Dieter},
  booktitle={Proceedings of The 6th Conference on Robot Learning},
  pages={785--799},
  year={2023}
}

@article{hu2024vpp,
  title={Video Prediction Policy: A Generalist Robot Policy with Predictive Visual Representations},
  author={Hu, Yucheng and Guo, Yanjiang and Wang, Pengchao and Chen, Xiaoyu and Wang, Yen-Jen and Zhang, Jianke and Sreenath, Koushil and Lu, Chaochao and Chen, Jianyu},
  journal={arXiv preprint arXiv:2412.14803},
  year={2024}
}

@article{openvla,
    title={OpenVLA: An Open-Source Vision-Language-Action Model},
    author={{Moo Jin} Kim and Karl Pertsch and Siddharth Karamcheti and Ted Xiao and Ashwin Balakrishna and Suraj Nair and Rafael Rafailov and Ethan Foster and Grace Lam and Pannag Sanketi and Quan Vuong and Thomas Kollar and Benjamin Burchfiel and Russ Tedrake and Dorsa Sadigh and Sergey Levine and Percy Liang and Chelsea Finn},
    journal = {arXiv preprint arXiv:2406.09246},
    year={2024}
}

@article{pi0,
  title   = {{$\pi_{0}$}: A {Vision-Language-Action} Flow Model for General Robot Control},
  author  = {{Physical Intelligence} and Kevin Black and Noah Brown and Danny Driess and Adnan Esmail and Michael Equi and Chelsea Finn and Niccolo Fusai and Lachy Groom and Karol Hausman and Brian Ichter and others},
  journal = {arXiv preprint arXiv:2410.24164},
  year    = {2024}
}

@article{pi05,
  title   = {{$\pi_{0.5}$}: A {Vision-Language-Action} Model with Open-World Generalization},
  author  = {{Physical Intelligence} and Kevin Black and Noah Brown and James Darpinian and Karan Dhabalia and Danny Driess and Adnan Esmail and Michael Equi and Chelsea Finn and Niccolo Fusai and others},
  journal = {arXiv preprint arXiv:2504.16054},
  year    = {2025}
}

@misc{3dgs,
      title={Towards Physically Executable 3D Gaussian for Embodied Navigation}, 
      author={Bingchen Miao and Rong Wei and Zhiqi Ge and Xiaoquan Sun and Shiqi Gao and Jingzhe Zhu and Renhan Wang and Siliang Tang and Jun Xiao and Rui Tang and Juncheng Li},
      year={2025},
      eprint={2510.21307},
      archivePrefix={arXiv},
      primaryClass={cs.CV},
      url={https://arxiv.org/abs/2510.21307}, 
}

@article{xVLA,
  title   = {X-VLA: Soft-Prompted Transformer as Scalable Cross-Embodiment Vision-Language-Action Model},
  author  = {Zheng, Jinliang and Li, Jianxiong and Wang, Zhihao and Liu, Dongxiu and Kang, Xirui
             and Feng, Yuchun and Zheng, Yinan and Zou, Jiayin and Chen, Yilun and Zeng, Jia and others},
  journal = {arXiv preprint arXiv:2510.10274},
  year    = {2025}
}

@inproceedings{rt2,
  title={RT-2: Vision-Language-Action Models Transfer Web Knowledge to Robotic Control},
  author={Zitkovich, Brianna and Joshi, Nikhil and Brohan, Anthony and Brown, Noah and Carbajal, Justice and Chebotar, Yevgen and Chen, Xi and Choromanski, Krzysztof and Ding, Tianli and Driess, Danny and others},
  booktitle={Proceedings of The 7th Conference on Robot Learning},
  pages={2165--2183},
  year={2023}
}

@inproceedings{palme,
  title={PaLM-E: An Embodied Multimodal Language Model},
  author={Driess, Danny and Xia, Fei and Sajjadi, Mehdi S. M. and Lynch, Corey and Chowdhery, Aakanksha and Ichter, Brian and Wahid, Ayzaan and Tompson, Jonathan and Vuong, Quan and Yu, Tianhe and others},
  booktitle={Proceedings of the 40th International Conference on Machine Learning},
  year={2023}
}

\end{document}